\documentclass[runningheads]{llncs}
\usepackage{graphicx}
\usepackage{tikz}
\usepackage{comment} 
\usepackage{amsmath,amssymb} % define this before the line numbering.
\usepackage{color}
\usepackage{multicol}
\usepackage{multirow}
\usepackage{booktabs}
\usepackage{algorithm}
\usepackage[noend]{algpseudocode}

\usepackage{caption}
\usepackage{subcaption}
\usepackage{url}

\expandafter\def\expandafter\UrlBreaks\expandafter{\UrlBreaks%  save the current one
  \do\a\do\b\do\c\do\d\do\e\do\f\do\g\do\h\do\i\do\j%
  \do\k\do\l\do\m\do\n\do\o\do\p\do\q\do\r\do\s\do\t%
  \do\u\do\v\do\w\do\x\do\y\do\z\do\A\do\B\do\C\do\D%
  \do\E\do\F\do\G\do\H\do\I\do\J\do\K\do\L\do\M\do\N%
  \do\O\do\P\do\Q\do\R\do\S\do\T\do\U\do\V\do\W\do\X%
  \do\Y\do\Z}

\begin{document}
% \linenumbers
\pagestyle{headings}
\mainmatter
\def\ECCVSubNumber{6254}  % Insert your submission number here

\title{PROFIT: A Novel Training Method for sub-4-bit MobileNet Models} % Replace with your title

\begin{comment}
\titlerunning{ECCV-20 submission ID \ECCVSubNumber} 
\authorrunning{ECCV-20 submission ID \ECCVSubNumber} 
\author{Anonymous ECCV submission}
\institute{Paper ID \ECCVSubNumber}
\end{comment}
%******************

%\begin{comment}
\titlerunning{PROFIT: A Novel Training Method for sub-4-bit MobileNet Models}
\author{Eunhyeok Park\inst{1}\orcidID{0000-0002-7331-9819} \and
Sungjoo Yoo\inst{2}\orcidID{0000-0002-5853-0675}}
\authorrunning{Eunhyeok Park and Sungjoo Yoo}
\institute{
\email{canusglow@gmail.com}~and~  \inst{2}\email{sungjoo.yoo@gmail.com}\\
\inst{1} Inter-university Semiconductor Research Center (ISRC) \\
\inst{2} Department of Computer Science and Engineering \and Neural Processing Research Center (NPRC)\\
\inst{1,2}Seoul National University, Seoul, Korea
}

%\end{comment}
%******************
\maketitle

\begin{abstract}
4-bit and lower precision mobile models are required due to the ever-increasing demand for better energy efficiency in mobile devices. In this work, we report that the activation instability induced by weight quantization (AIWQ) is the key obstacle to sub-4-bit quantization of mobile networks. To alleviate the AIWQ problem, we propose a novel training method called PROgressive-Freezing Iterative Training (PROFIT), which attempts to freeze layers whose weights are affected by the instability problem stronger than the other layers. We also propose a differentiable and unified quantization method (DuQ) and a negative padding idea to support asymmetric activation functions such as h-swish. We evaluate the proposed methods by quantizing MobileNet-v1, v2, and v3 on ImageNet and report that 4-bit quantization offers comparable (within 1.48~\% top-1 accuracy) accuracy to full precision baseline. In the ablation study of the 3-bit quantization of MobileNet-v3, our proposed method outperforms the state-of-the-art method by a large margin, 12.86~\% of top-1 accuracy. The quantized model and source code is available at \url{https://github.com/EunhyeokPark/PROFIT}.
\end{abstract}

\section{Introduction}

Neural networks are widely adopted in various embedded applications, e.g., smartphones, AR/VR devices, and drones. 
Such applications are characterized by stringent constraints in latency (for real-time constraints) and energy consumption (because of battery).
Thus, it is imperative to optimize neural networks in terms of latency and energy consumption, while maintaining the quality of the neural networks, e.g., accuracy.

Quantization is one of the most effective optimization techniques.
% 여기에 industry & academic reference들 다 추가.
By reducing the bit-width of activations and weights, 
the performance and energy efficiency can be improved by executing more computations with the same amount of memory accesses and computation resources (e.g., the silicon chip area and battery). The 8-bit computation is already popular% and contributes to wide-spread applications of neural networks
~\cite{SNAP,song20197,IntArithmetic,NVIDIA_8bit}, and NVIDIA recently announced that tensor core supports 4-bit precision which gives more than 50~\% of performance improvement on ResNet-50~\cite{NVIDIA_4bit}. We expect 4-bit and lower precision computation will become more and more important and make further contributions to the energy efficiency and real-time characteristics of future deep learning applications~\cite{OLAccel,Samsung_NPU,BitFusion,Andrew,IntArithmetic}.

\begin{figure}[t!]
    \centering
  \includegraphics[width=0.8\columnwidth]{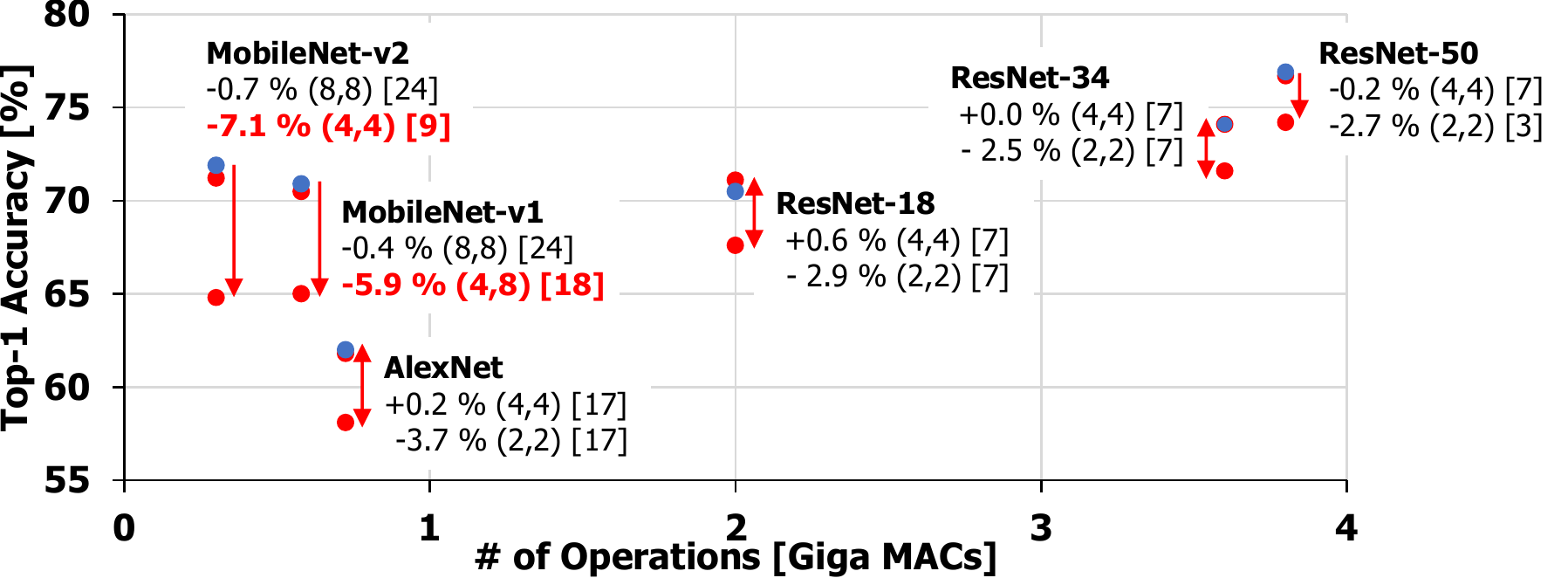}
  \caption{Recent results of  quantization studies. The blue dots represent full-precision accuracy and the red dots quantized accuracy. The tuple (a,w) represents the bit-width of activation and weight, respectively. } 
  \label{img:overview}
\end{figure}

% 은혁, 그림 1이 좀 애매해. 한눈에 들어오지 않고. 메시지가 그림으로는 안보임. Text를 보면 옛날에 개발된 네트워크 양자화는 좀 되나, optimized network에서의 양자화는 아직 기존 기술로 4비트까지는 어렵다는 것 보여주는 것 같은데, text 안보고 그림만 봐도 그걸 알 수 있어야 하거든. 보통 논문 읽는 사람들이 1번 그림은 self-contained 된 것 기대하는 경향 있어. 그렇다고 그림의 caption에 주절이 주절이 설명달면 그것도 안좋고.
% 그리고, 각 모델의 대표 point가 정확히 어떤 precision에서의 값인지가 명확하지 않아. 그림 자세히 보고 추측하면 각 모델에서 제시하는 8비트 이하 정확도 중 가장 높은 걸 보여주는 것 같은데, 그 경우도 GFlops는 어떤 precision 경우의 것인지 모르겠어. 물론 Flops니까 32비트일꺼라고 추측은 되는데, 이것도 명시되어 있지 않고.
% 아마 시도 했을 것 같은데, 내 생각에는 옛날 모델들은 4비트 경우 best 하나씩만 보여줘서 단순화 하고. 그때의 GBOPS로 표현하고. 아니면 좀더 직접적으로 4비트 operation 갯수 기준으로 표현할 수도 있을 듯.
% MobileNet v1,2,3 경우를 현재 모은 기존 결과들 top1 vs GBOPS 를 보여주면 어떨까 해.

% 성능 및 model size에 대한 비교가 실험에 있어서 여기서는 motivation을 강조하고 싶었습니다. operation 숫자가 적은 model에 양자화가 안된다는 점을 강조하려 했습니다. 좀 더 간단하게 정리해서 그림을 다시 그려봤습니다. 

In order to support the up-coming hardware acceleration, there have been active studies on sub-4-bit quantization~\cite{PACT,PACT_only,BNN_Hubara,QIL,BiReal,XNOR,Dorefa,Balanced,NVIDIA_Ternary,CVPR}. % and the implementations on CPU, GPU~\cite{Andrew, IntArithmetic}, and dedicated hardware accelerators~\cite{OLAccel,BitFusion}.  
These studies show that deep networks, e.g., AlexNet or ResNet-18 for ImageNet classification~\cite{ImageNet}, can be quantized into sub-4 bits with negligible accuracy loss, as shown in Figure \ref{img:overview}. However, these out-dated networks are prohibitively expensive to use in mobile devices. In mobile devices, it is imperative to quantize the optimized networks, e.g., MobileNet-v2~\cite{V2} or MobileNet-v3~\cite{V3}.% in low precision.

However, the previous quantization methods do not work well on the advanced optimized networks. These networks adopt novel structures like depth-wise separable convolution~\cite{V1}, inverted residual block with linear expansion layer~\cite{V2}, squeeze-excitation module, and h-swish activation function~\cite{V3}. These structures have less redundancy and are vulnerable to quantization, and the h-swish activation function generates an asymmetric distribution. Previous quantization methods did not consider the optimizations, thus having a significant accuracy degradation in the sub-4-bit quantization of the advanced networks.

In this study, we propose two novel ideas that enable 4-bit quantization for the optimized networks. First, we report that the primary reason for the accuracy loss in sub-4-bit quantization is the activation instability induced by weight quantization (AIWQ). Weight quantization can skew the statistics of the output activation, i.e., the mean and variance, at every iteration during fine-tuning. This adversely affects the following layers and finally prevents the network from converging to a good optimal point in low-precision quantization. In order to address this problem, we propose a novel training method called \textbf{PRO}gressively-\textbf{F}reezing \textbf{I}terative \textbf{T}raining (PROFIT) that minimizes the effects of AIWQ by progressively freezing the weights sensitive to AIWQ during training.

%In training time, we obtain running mean and variance and, in test time, use them in the batch normalization layer. Our analysis shows that the running mean and variance are hurt by the activation instability induced by weight quantization (AIWQ). 

%Fine-tuning is essential to recover from the accuracy loss of sub-4-bit quantization~\cite{PACT,QIL}. However, the fine-tuning tends to increase the frequency of weights near the quantization thresholds. In such a case, a small change of the full-precision weight (by back-propagation) near the quantization threshold is amplified by the weight quantization, i.e., the rounding operation. This causes the convolution output to significantly change across batches, which we call the AIWQ problem. 

%The activation instability perturbs the activation distribution, which finally prevents us from obtaining good running mean and variance in training time. Such poor running mean and variance make batch normalization less effective in test time.  
%This instability also hinders with the convergence of weight, which leads to the degradation of test accuracy. 실험으로 명확히 보여주진 못한 얘기라 일단 제외.
%In order to address this problem, we propose a novel training method called BLast (BN Last) that tries to minimize the effects of AIWQ by judiciously performing the training of batch normalization (BN) layer while progressively freezing the weights sensitive to the AIWQ problem.

Second, we identify the limitations of the state-of-the-art trainable methods of linear quantization in terms of asymmetric activation support, and we propose a novel quantization method called differentiable and unified quantization (DuQ) and negative padding. Many advanced networks begin to adopt the activation functions allowing a small amount of the negative value, e.g., hard swish (h-swish) of MobileNet-v3 and Gaussian error linear unit (GeLU) of BERT~\cite{Bert}. These activation functions increase accuracy with minimal overhead. However, they produce asymmetric output distributions. Existing quantization methods are only designed to support symmetric or non-negative distributions; therefore, they are unsuitable for the asymmetric distributions. The proposed DuQ method resolves the above problems without limiting the value range while minimizing both rounding and truncation errors in a differentiable manner. Furthermore, the novel negative padding idea contributes to accuracy improvement by appropriately utilizing the quantization levels under an asymmetric distribution.

%The state-of-the-art methods try to minimize either rounding error~\cite{QIL} or truncation error~\cite{PACT}. It is desired to minimize both errors for further reduction in precision.
%In addition, existing methods like PACT and QIL support only a limited value range of output activation, e.g., non-negative or [0,1]. Thus, they cannot be applied to novel activation functions, e.g., h-swish function, which utilizes negative activations and novel structures like squeeze-and-excitation module and expansion layer which both require producing both positive and negative activations. Our proposed DuQ method resolves the above problems without limiting the value range while minimizing both rounding and truncation errors in a differentiable manner. 

% As will be shown in Figure \ref{img:perf_model}, our proposed methods enable 4-bit quantization of optimized networks at high accuracy thereby pushing mobile networks towards a more resource-efficient regime compared with the state-of-the-art quantization solutions.

% By combining Blast and DuQ, we can quantize the optimized network with minimal accuracy loss. ...

%-------------------------------------------------------------------------
\section{Related Work}
The neural network architecture has been continuously improved while increasing accuracy at a lower computation cost. %MobileNet-v1 introduced depth-wise separable convolutions to reduce the computation cost~\cite{V1}. % of the convolution layer~\cite{V1}. 
%MobileNet-v2 proposed an inverted residual structure where, unlike in the  bottleneck structure, the input is expanded for more comprehensive representation ability~\cite{V2}.% thanks to the low cost depth-wise separable convolution~\cite{V2}. 
MobileNet-v1~\cite{V1} and -v2~\cite{V2} introduced a depth-wise separable convolution and an inverted residual structure respectably.
MNasNet was designed based on AutoML, which automates the network architecture search, considering the computation cost~\cite{MNasNet}. MobileNet-v3 is the state-of-the-art network, which was designed from MNasNet by improving it with the h-swish activation function and squeeze-excitation modules~\cite{V3}. % that take the global averaged activation and generate channel-wise scaling~\cite{V3}.
Compared to the conventional deep networks like AlexNet~\cite{AlexNet}, VGG~\cite{VGG}, and ResNet~\cite{ResNet}, the recent networks have adopted optimized structures for better accuracy at a low computation cost. However, such optimized structures render quantization challenging, especially for 4-bit and lower precision quantization.

Several studies have been proposed for sub-4-bit quantization. \cite{XNOR} and \cite{Dorefa} are the representative studies showing that neural networks can be quantized into sub-4-bits with marginal accuracy loss. \cite{CVPR} proposed progressive quantization, which reduces the precision in a progressive manner. \cite{PACT,PACT_only,LSQ,DSQ,QIL} and \cite{WRPN} show that networks for large-scale datasets, e.g., ResNet~\cite{ResNet} for ImageNet~\cite{ImageNet}, can be quantized into sub-4-bits without accuracy loss. Recently, \cite{FightBias} and \cite{DataFree} are focused on post-training 8-bit quantization, and showed that quantization introduces a bias shift, which acts as the main cause of accuracy degradation. %However, we pursue 4-bit and lower precision quantization, and we focus on the effects of weight quantization during training. 
The previous works show the potential of aggressive quantization in terms of sub-4-bits (with fine-tuning) for AlexNet and ResNet, or 8-bit quantization (without fine-tuning) for MobileNet-v2. %However, the target networks, e.g. AlexNet and ResNet, are not optimized. 
In this study, we focus on the 4-bit and lower precision quantization (with fine-tuning) for recently optimized networks such as MobileNet-v2 and -v3. 

Many of the previous quantization methods utilize hand-crafted loss, e.g., L2 distance between full-precision and quantized data~\cite{PACT}. 
For a lower precision, it is desirable to learn the quantization interval of the loss of the target task, e.g., cross-entropy loss for classification. In \cite{PACT,PACT_only,QIL} and \cite{LSQ}, the quantization interval is learned via backpropagation. In this study, we also propose DuQ with a negative padding idea that learns the quantization interval by utilizing the gradients and asymmetric activations better than the previous methods. %Learnable step-size quantization (LSQ)~\cite{LSQ} is another differentiable quantization algorithm. This could be extended to assymmetric quantization, but it handles only symmetric or non-negative distributions. 
%In section \ref{pact_qil}, we will provide more detailed analyses of existing quantization algorithms and explain how our proposed DuQ and negative padding methods improve on them. 

\section{AIWQ and PROFIT}
In this section, we first demonstrate that the AIWQ problem is strongly correlated with the accuracy degradation at low precision. Then, we present a metric to measure the activation instability and propose a training method called PROFIT that controls the training of each layer to minimize the effect of AIWQ based on the presented metric. %We also propose a training method called BLast which, based on the activation instability metric, controls the training of each layer to minimize the effect of activation instability thereby offering better test accuracy. 

\begin{figure*}[t]
  \centering
  \includegraphics[width=\columnwidth]{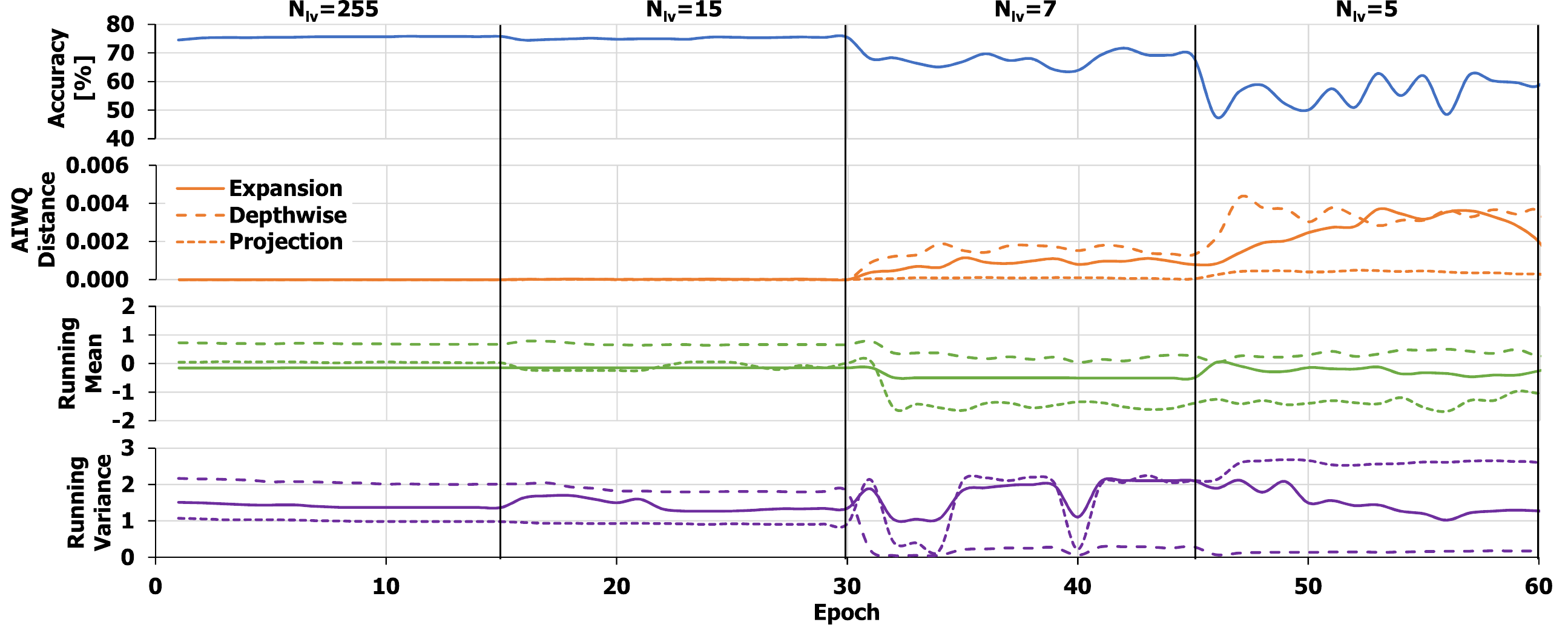}
  \caption{ Top-1 accuracy [\%], AIWQ distance, and the running mean, and the running variance during fine-tuning for quantization where $N_{lv}$ represents the number of available quantization levels. Running mean and variance are extracted from the arbitrary channels of batch normalization layer after the depth-wise convolution in the 2nd inverted residual module of MobileNet-v3.  }
  \label{img:ft_curve}
\end{figure*}

% 은혁, 이 그림의 Bottleneck --> Reduction 으로 변경해야 할 듯.
% 그리고, 이 그림의 경우는 Expansion layer가 큰 AIWQ를 보이는데, 그림 4에서는 AIWQ 값이 작아. 비록 이 그림은 Cifar100이고 그림 4는 ImageNet 데이터이지만 그래도 reviewer가 궁금해 할 것 같아. 최소한 dataset에 따라 per-layer AIWQ가 많이 차이 나는가?는 질문 던질 수 있을 듯.

%\subsection{Notation}
%First, we explain the notation used in the paper. The subscripts $l, i$, and $o$ represent the layer index, input channel index, and output channel index, respectively, and $W, I$, and $O$ represent weight, input activation, and output activation, respectively. In addition, the superscript $t$ denotes the iteration index. For instance, $W^t_{l,o,i}$ represents the weights updated at $t$-th iteration on $l$-th layer, $o$-th output channel and $i$-th input channel. In the case of weight-quantized convolution layer, the output activation can be expressed as follows:
%\begin{equation}
%O^t_{l,o} = \sum_i Q(W^t_{l,o,i}) \otimes I^t_{l,i},
%\end{equation} 
%where $\otimes$ is the convolution operation and $Q$ is the quantization function. 
%Note that, for simplicity, activation quantization, which is performed after BN layer, is not shown in the equation.

\subsection{Observation} \label{aiwq}
Figure \ref{img:ft_curve} (Accuracy) shows the accuracy of MobileNet-v3 for the Cifar-100 dataset\\~\cite{CIFAR}. The accuracy is measured during fine-tuning in progressive weight quantization~\cite{CVPR}, where the number of quantization levels of the weights $N_{lv}$ are gradually reduced from 255 to 5 while using full-precision activation. %The orange line, denoted by Training represents the accuracy obtained by using the mean and variance of the current training batch at the batch normalization layers. Meanwhile, the blue line, denoted by Test represents the accuracy measured by using the running mean and variance which are obtained during training, i.e., continuously updated at each batch for later test time usage.
The accuracy curves in Figure \ref{img:ft_curve} show that, in each precision case, e.g., $N_{lv}=15$, the test accuracy is gradually recovered as the fine-tuning advances. However, when the number of available quantization levels reduces to 7 or lower, the accuracy significantly oscillates and fails to converge. 

\begin{figure}[t!]
    \centering
  \includegraphics[width=0.8\columnwidth]{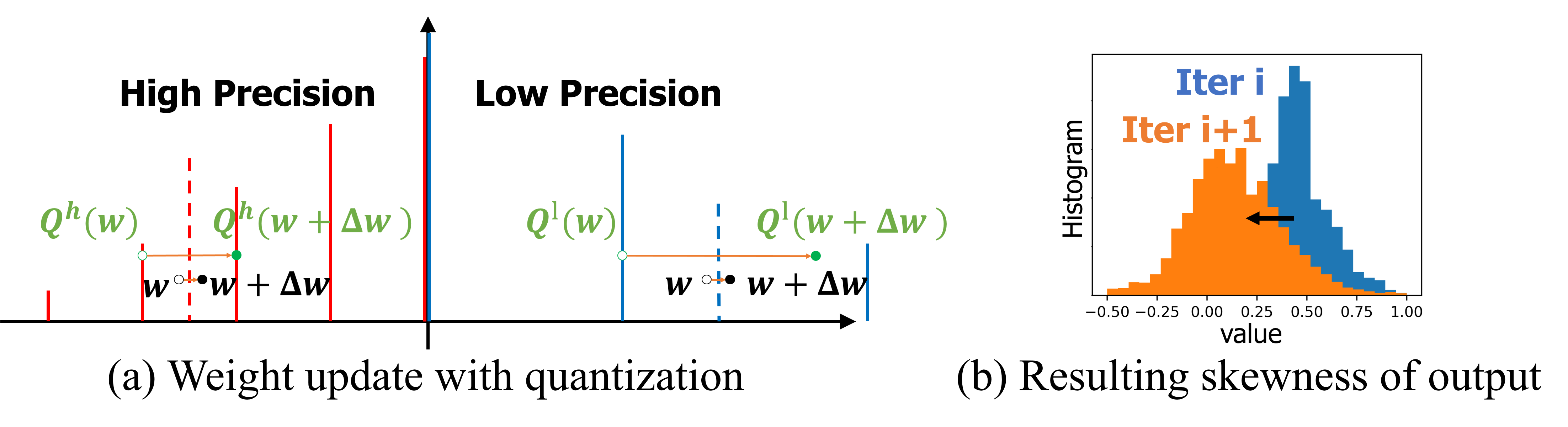}
  \caption{Activation instability induced by weight quantization.} 
  \label{img:ch_hist}
\end{figure}

From our analysis, that will be given in the next subsection, this is mainly due to the activation instability during fine-tuning, as shown in Figure \ref{img:ch_hist}. The challenge is that the effects of the weight update, due to backpropagation, could be amplified by the quantization operation, i.e., rounding operation.
Particularly, as Figure \ref{img:ch_hist}~(a) shows, if a weight near the quantization threshold is updated to change its value crossing the threshold, then the quantization will result in different quantized weight values before and after the weight update. Thus, the results of the convolution operation will change due to the weight update and quantization. As Figure \ref{img:ch_hist} (b) shows, this skews the statistics of the output activation, which affects the following layers, including the normalization layers; therefore, yielding inaccurate running mean and variance in these layers. The inaccurate running mean and variance, obtained during training, degrades the test accuracy as illustrated in Figure \ref{img:ft_curve} because they are utilized in the normalization layers during the test but can't represent the actual statistics of the activation.

At lower precision, the activation instability induced by weight quantization (AIWQ) becomes more significant because the space between two adjacent quantization levels ($\sim$ 1/$2^\textbf{bit-width}$) becomes large. Thus, the lower the precision gets, the more the activation instability can be incurred.\footnote{Note that, in higher (lower) precision, there will be more (less) occurrences of smaller (larger) amounts of activation instability. Our study empirically shows that a few occurrences of large activation instability at low precision tend to have a higher impact on the test accuracy than many occurrences of small activation instabilities at high precision.} Besides, please note that the AIWQ problem is also found in conventional neural networks. However, because these networks use full convolution as their building block having more than hundreds of accumulation per output, the quantization error is likely to be amortized based on the law of large numbers. This makes the networks robust to quantize, but they also suffer from instability when the precision lower.

%The perturbed activation prevents us from convergence to good minima as demonstrated in Figure \ref{img:ft_curve}. Note that the effects of perturbed activation on the next layer tends to be small due to the batch normalization layer. Only the running mean and variance of the normalization layer are mainly affected by the activation instability since they are obtained by exponential moving average of mean and variance of the un-normalized activation. This is why the test accuracy in Figure \ref{img:ft_curve} heavily oscillates while the training accuracy gives much smaller oscillations. 
%In the next subsection, in order to analyzes AIWQ in detail, we propose a metric to measure the magnitude of AIWQ.

% 가장 큰 에러를 제거하는 방법은 

%이 중 최종 정확도에 더 크게 영향을 미치는 것은 batch normalization layer의 running mean & variance 이다. 그림 2의 2-bit training/test mode accuracy curve를 보면 test mode는 심하게 요동치는 반면 training curve는 상대적으로 훨씬 적게 요동치며 학습이 꾸준히 진행되는 것을 확인할 수 있다. fine-tuning 도중에는 batch normalization의 경우 실제 데이터의 mean & variance를 구해 normalize하므로 WQAF가 심해지더라도 이를 보정하는 효과가 있어 상대적으로 다음 layer에 전달되는 에러를 보정해준다. 그러나 running mean과 variance의 경우 변화하는 mean과 variance를 exponential moving average로 추적하여 구하게 되는데 이때 에러가 계속 누적되게 되므로 에러의 영향을 크게 받게된다.  <- 수식을 동원하여 정리해보자
% 이러한 특성은 training curve가 꾸준히 증가하는 점에서도 확인할 수 있다. 

% TODO: 수학적으로 표현을 바탕으로 설명을 가다듬기 

\subsection{Activation Instability Metric}
We present a metric to quantify the per-layer activation instability and use it to (1) prove that AIWQ is correlated with the test accuracy of the low-precision model (in Figure \ref{img:ft_curve}) and (2) utilize the per-layer sensitivity when determining the order of freezing the weights during training (to be explained in Section \ref{sec_blast}).

In order to measure per-layer activation instability, a desirable solution would be to calculate the KL divergence between two distributions of the outputs before ($p^t_o$) and after ($q^t_o$) a training iteration $t$. We first calculate the per-output channel KL divergence between $p^t_o$ and $q^t_o$. Then, as shown below, we compute the layer-wise AIWQ metric $D^l$ by averaging the per-output channel KL divergence across all the output channels of the current layer in the current training batch. 
\begin{equation}
    D^l = E_o^t\big[D_{KL}(p^t_o ~||~ q^t_o)\big].
    \label{aiwq_metric}
\end{equation}
\begin{figure}[t!]
    \centering
  \includegraphics[width=\columnwidth]{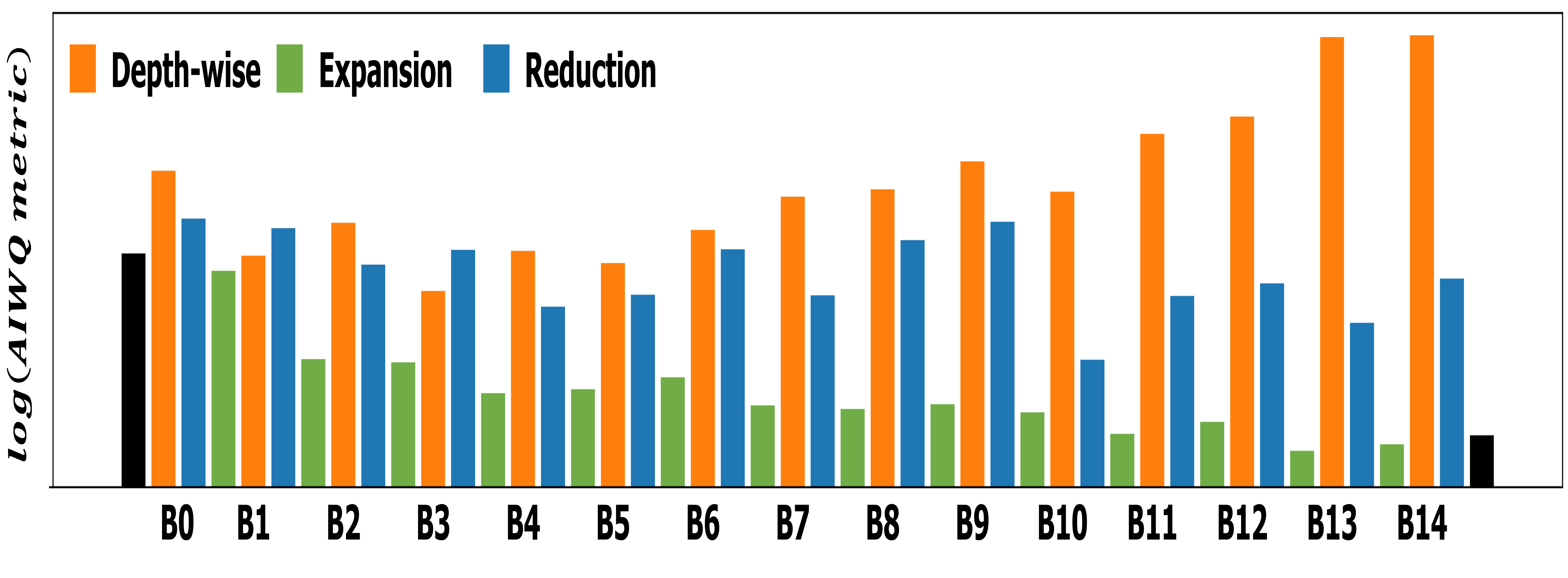}
  \caption{Layer-wise AIWQ metric measured at MobileNet-V3 on Imagenet. Bx refers to x-th building block of MobileNet-v3. Please note that y-axis is in log-scale thus the layer-wise sensitivity is highly vary depending on layer index. } 
  \label{img:layerwise}
\end{figure}

We simplify the computation of the KL divergence by adopting a second-order model which considers the mean and variance of per-channel output distribution because the computation of the KL divergence is expensive, and the second-order model proves sufficient for our goal of evaluating the relative order between layers.\footnote{
Note that in ~\cite{DataFree,FightBias} the first-order momentum, i.e., the channel-wise mean is utilized to evaluate the difference in the output distributions. From our observation, the channel-wise variance is also significantly skewed by the AIWQ,  and significantly affects the accuracy via normalization. AIWQ metric is designed to consider the two important momenta, the mean and variance, concurrently with an affordable cost.}
Figure \ref{img:layerwise} shows the AIWQ metric of MobileNet-v3 on ImageNet. 
As the figure shows, depth-wise convolution layers tend to exhibit a large AIWQ while some reduction (point-wise 1x1 convolution) layers also give a larger AIWQ than the depth-wise layers in the early layers.

%The simplified process as follows. First, we approximate the output of weight-quantized convolution before and after weight update as uni-variate Gaussian random variables. 
%\begin{align}
%    p^t_o \approx N\bigg(\mu_o, \sigma_o~||~\sum_i Q(W^t_{l,o,i}) \otimes I^t_{l,i}\bigg), \\
%    q^t_o \approx N\bigg(\mu'_o, \sigma'_o ~||~\sum_i Q(W^{t-1}_{l,o,i}) \otimes I^t_{l,i}\bigg).
%\end{align}
%where $p^t_o$ and $q^t_o$ represent the approximate distribution of convolution output on a channel after (at iteration index $t$) and before (at index $t-1$) weight update, respectively.
%Note that the same input activation $I^t_{l,i}$ is used to evaluate the effect of weight quantization on the convolution output.

% The AIWQ varies during fine-tuning. Recall Figure \ref{img:ft_curve}, which illustrates how the AIWQ metric varies during fine-tuning.% in the case of the 2nd inverted residual block of MobileNet-v3 on Cifar100. 
Recall Figure \ref{img:ft_curve}, which illustrates how the AIWQ metric varies during fine-tuning. The AIWQ increases in the 7- and 5-level quantization, which empirically proves that weight quantization at low precision can incur significant perturbation of the convolution output, i.e., makes the output activation unstable. 
As shown in the figure, such an instability causes the running mean and variance\footnote{We show the mean and variance on three sampled output channels in Figure \ref{img:ft_curve}.} to fluctuate, which prevents us from obtaining good running mean and variance during training.
When comparing the accuracy, the AIWQ metric, and the mean and variance in Figure \ref{img:ft_curve}, they are closely correlated at low precision, i.e., $N_{lv}= 5$.
In the following subsection, we use the AIWQ metric to schedule which layers to freeze first during the fine-tuning, which contributes to reducing the AIWQ, hence improving the test accuracy.

\subsection{PROFIT}
\label{sec_blast}

%The running mean and variance affected by AIWQ can be corrected by obtaining statics after training. However, AIWQ also affects the weights of following layers thus we proposed a training pipeline aiming minimization of AIWQ during training. 

%As shown previously, the running mean and variance are affected by AIWQ. Our basic idea is to additionally train the batch normalization and quantization layer, after freezing all the other weights in the last training step of training, which is called BN last training, in short, BLast.
%In the last training step, there is no AIWQ since the weights are freezed. Thus, the additional training can offer better running mean and variance. 

\begin{algorithm}[t]
\caption{Pseudo code of PROFIT algorithm}\label{alg:profit}
\begin{algorithmic}[1]
\State Initialize network and full-precision training (+ progressive quantization)
\State Set quantization parameters according to the target bit-width.

\Procedure{AIWQ sampling}{}
\For{layer $\in$ network.layers}
\If{layer is quantized convolution}
    \State metric\_map[layer] = 0
\EndIf
\EndFor
\For{i $\in$ sampling~iterations}
    \For{layer $\in$ network.layers}
        \State layer.forward()
        \If{layer is quantized convolution}
            \State metric\_map[layer] += layer.AIWQ\_metric
        \EndIf
    \EndFor
\State network.update()
\EndFor
\EndProcedure

\State AIWQ\_list $\gets$ sort\_by\_value(metric\_map, order=descending)
\State $N_{layers}$ $\gets$ len(AIWQ\_list)

\Procedure{PROFIT}{}
\For{n $\in$  $N_{PROFIT}$}
    \For{e $\in$ Profit\_Epoch}
        \State network.training\_epoch()
    \EndFor
    
    \State begin $\gets n * N_{layers} /N_{PROFIT}$, end $\gets (n+1) * N_{layers} /N_{PROFIT}$
    \State freeze\_target\_layers $\gets$ AIWQ\_list[begin:end]
    
    \For{layer $\in$ freeze\_target\_layers}
        \State layer.learning\_rate $\gets$ 0
    \EndFor
\EndFor

\For{e $\in$ BN\_Epoch}
        \State network.training\_epoch()
    \EndFor
    
\EndProcedure
%\Procedure{MyProcedure}{}
%\State $\textit{stringlen} \gets \text{length of }\textit{string}$
%\State $i \gets \textit{patlen}$
%\BState \emph{top}:
%\If {$i > \textit{stringlen}$} \Return false
%\EndIf
%\State $j \gets \textit{patlen}$
%\BState \emph{loop}:
%\If {$\textit{string}(i) = \textit{path}(j)$}
%\State $j \gets j-1$.
%\State $i \gets i-1$.
%\State \textbf{goto} \emph{loop}.
%\State \textbf{close};
%\EndIf
%\State $i \gets i+\max(\textit{delta}_1(\textit{string}(i)),\textit{delta}_2(j))$.
%\State \textbf{goto} \emph{top}.
%\EndProcedure
\end{algorithmic}
\end{algorithm}

%\begin{figure}[ht!]
%    \centering
%  \includegraphics[width=0.5\columnwidth]{eccv2020kit/figure/blast.pdf}
%  \caption{PROFIT method.} 
%  \label{img:blast}
%\end{figure}

We propose a novel training method which aims at minimizing the AIWQ effect to improve the accuracy of low precision networks. Our basic idea is progressively freezing (the weights of) the most sensitive layer to AIWQ to remove the fluctuation source, thus allowing the rest of the layers to converge to a more optimal point. We determine the layer-wise order of the weight freezing considering the per-layer AIWQ metric in Eqn. \ref{aiwq_metric}. 
Algorithm \ref{alg:profit} shows how our method, called \textbf{PRO}gressively-\textbf{F}reezing \textbf{I}terative \textbf{T}raining (PROFIT), works. When PROFIT is triggered, we start a sampling stage where we evaluate the per-layer AIWQ metric for each layer. 
After the sampling stage, we perform fine-tuning in an initial stage without freezing weights.
Subsequently, after sorting all the weight layers in terms of the per-layer metric, we perform weight freezing stages by selecting the most sensitive layers (the ones having the largest AIWQ metric values) and freezing their weights. As shown in the algorithm, we iteratively perform $N_{PROFIT}$ freezing stages. Thus, in each stage, $N_{layers}$/$N_{PROFIT}$ ($N_{layers}$ is the total number of quantized layers) are selected from the sorted layer list and their weights are frozen. Then, we perform training for all un-frozen layers. After finishing the additional training stage, we select the next set of un-frozen sensitive layers (another $N_{layers}$/$N_{PROFIT}$ layers) and repeat the same procedure until there is no more un-frozen layer left. Finally, we perform an additional training stage (typically, 3-5 epochs) for the normalization layers while freezing all the other layers. This further stabilizes the statics of the normalization layers. As will be shown in our experimental results, PROFIT improves the accuracy of low precision networks by significantly reducing the effect of AIWQ. 

%BLast+ helps to reduce AIWQ noise for not only running mean and variance but also the weights of subsequent layers. 

% 아래 부분 실험 결과 없어서 일단 제외함
% Note that, as shown in Figure \ref{img:duq}, 
% we decrease the learning rate in each stage. 
% In our experiments, we will separately report the effects of 
% learning rate control and BN last training.

% WQAF의 영향에 의해 BN parameter와 other parameter들이 영향을 받아 정확도가 감소한다. 이러한 정확도 감소를 복구하기 위하여 두 가지 학습 방법을 제시한다. 우선 가장 많은 부분을 차지하는 BN parameter의 문제는 학습이 끝난 후 weight를 고정한 상태에서 Batch Normalization layer만 학습을 진행함으로써 보정할 수 있다. 이 방법을 batch normalization last training (BLast)라 부른다. 그림에서 볼 수 있듯이, Blast phase에서 BN layer만 학습했을 때 Test accuracy가 빠르게 복구되는것을 확인할 수 있다. 그러나 그 외에 노이즈에 의해 다른 레이어에 미친 영향의 경우의 다른 방법으로 복구해야 한다. 이 논문에서는 앞선 방법에 대해 제안한 메트릭 기반으로 가장 민감한 레이어부터 순차적으로 고정하여 네트워크를 점차 수렴해가는 방법, BLast+를 제안한다. 

% 그림 X는 Blast+의 training schedule를 시각화한 것이다. Bit target이 변한 상태에서 처음 Quantization을 적용하고 나서 우선 Quantization parameter 및 batch normalization parameter만 학습시켜 quantization function에 의해 변화된 runing mean & variance에 네트워크를 적응시킨다. 그 후 sampling period 동안 Metric을 측정하여 layer 별 WQAF 영향을 측정하고 민감도에 따라 레이어를 분류한다. 그다음 Nth repeated training phase동안 민감한 순서대로의 N-th layer를 고정시키고 학습을 진행시키고 이를 반복함으로써 네트워크를 점차 좋은 포인트로 수렴시켜나간다. 우리의 관찰에 따르면 레이어의 sensitivity order는 일정하게 유지되므로, KL divergence는 앞 부분의 짧은 iteration에서만 측정하여 비용을 최소화한다. 이렇게 순차적으로 WQAF의 영향이 큰 레이어들을 고정시켜나감으로써 noise source를 점차 제거해가며 네트워크를 더 좋은 지점으로 수렴시킬 수 있게 된다. 

% 여기까지 봄

\section{Quantization for Asymmetric Distributions}

\begin{figure}[t!]
    \centering
  \includegraphics[width=0.9\columnwidth]{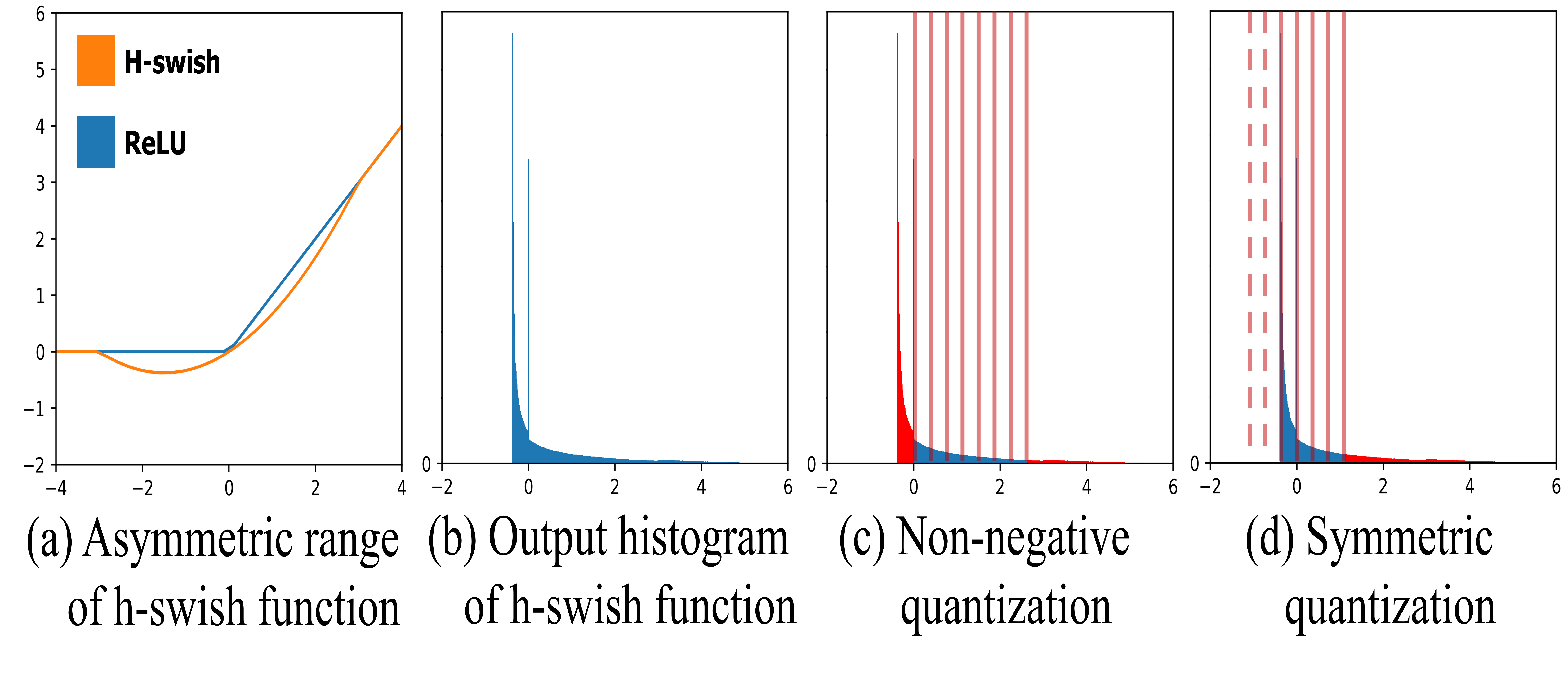}
  \caption{Characteristics of h-swish function and corresponding non-negative and symmetric quantization.} 
  \label{img:dist}
\end{figure}

In many advanced networks, the activation functions allow a small number of negative values, e.g., h-swish of MobileNet-v3 and GeLU of BERT, are becoming more popular. These functions increase the accuracy with minimal computation overhead, thus gradually expanding their scope of use. However, they have a critical limitation in terms of quantization. Because of the negative range, the output has an asymmetric distribution. Even though the negative range is small, many values are concentrated in that area, as shown in Figure \ref{img:dist}. These negative values should be carefully considered to maintain accuracy at low precision. 

However, existing quantization methods %, such as PACT~\cite{PACT}, QIL~\cite{QIL}, or LSQ~\cite{LSQ}, 
are only designed for symmetric or non-negative output.
%and also most of the existing hardware only supports symmetric (signed integer) or non-negative (unsigned integer) operation. 
In such a case, when we apply quantization to only the non-negative output of the h-swish function, many negative values are ignored (Figure \ref{img:dist}~(c)). On the contrary, when we apply symmetric quantization, some of the quantization levels, allocated for large negative values, are wasted, and a significant truncation error is incurred due to the narrower value range for positive values (Figure \ref{img:dist}~(d)). In either case, there is a significant loss in accuracy. 

In order to quantize the asymmetric distribution with minimal accuracy loss, we propose two ideas: DuQ and negative padding. First, we propose a quantization algorithm called Differentiable and Unified Quantization (DuQ) that resolves the above problems without limiting the value range while minimizing the rounding and truncation errors in a differentiable manner. Second, we propose negative padding that allows us to avoid wasting quantization levels, hence improving accuracy at low precision.

%\subsection{Rounding and Truncation Errors}
%Figure \ref{img:error} shows that quantization has two error sources, rounding and truncation. As shown in the figure, depending on whether a data is located inside/outside of quantization interval, rounding/truncation error is incurred. 
%Both errors are closely related with each other. For instance, the smaller truncation error (by increasing the truncation threshold) can incur the larger rounding error (due to the larger quantization interval).
%Thus, the quantization method needs to find the best trade-off between the two errors in minimizing the training loss.

%\begin{figure}[ht!]
%    \centering
%  \includegraphics[width=\columnwidth]{latex/figure/errors.pdf}
%  \caption{Two error sources of quantization.} 
%  \label{img:error}
%\end{figure}

\subsection{Limitations of State-of-the-Art Methods}
\label{pact_qil}
Our goal is to realize differentiable quantization, which minimizes the task loss of the asymmetric distribution of activation.
There are three representative methods, parameterized clipping activation function (PACT)~\cite{PACT}, quantization interval learning (QIL)~\cite{QIL}, and learned step size quantization (LSQ)~\cite{LSQ}. In all the methods, the differentiable parameters and quantization intervals are updated through backpropagation to minimize the task loss.

\begin{figure}[t]
    
    \centering
    \begin{subfigure}{.32\columnwidth}
        \centering
        \includegraphics[width=\columnwidth]{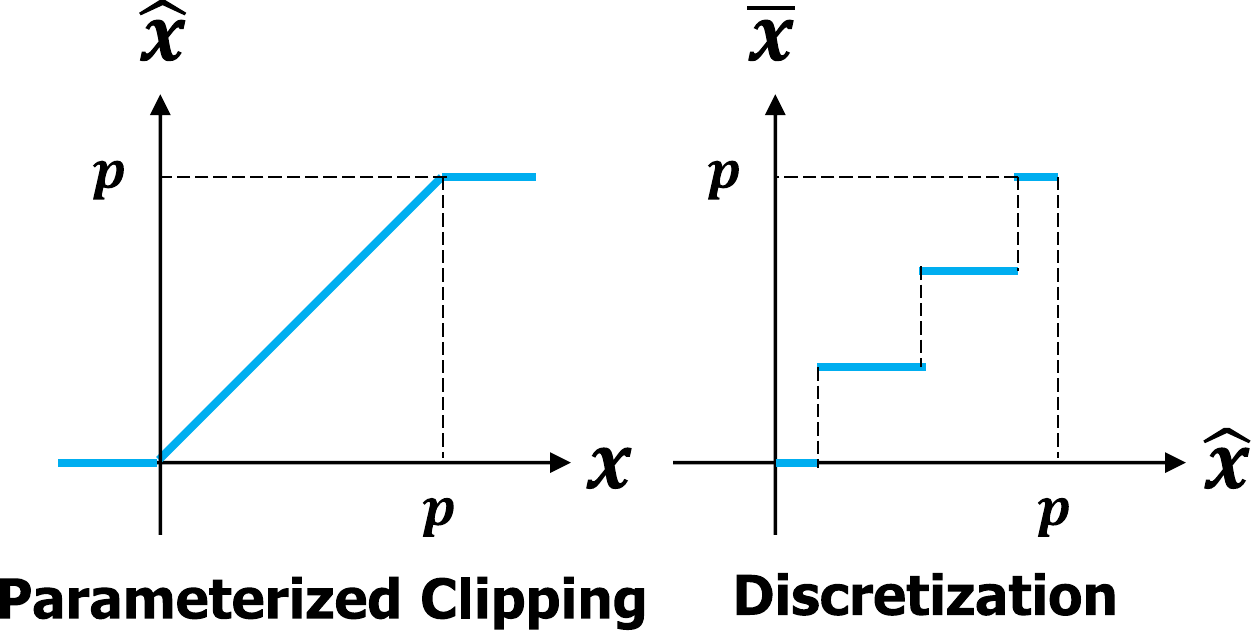}
        \caption{PACT algorithm.} 
        \label{img:pact}
    \end{subfigure}
    \begin{subfigure}{.32\columnwidth}
        \centering
        \includegraphics[width=\columnwidth]{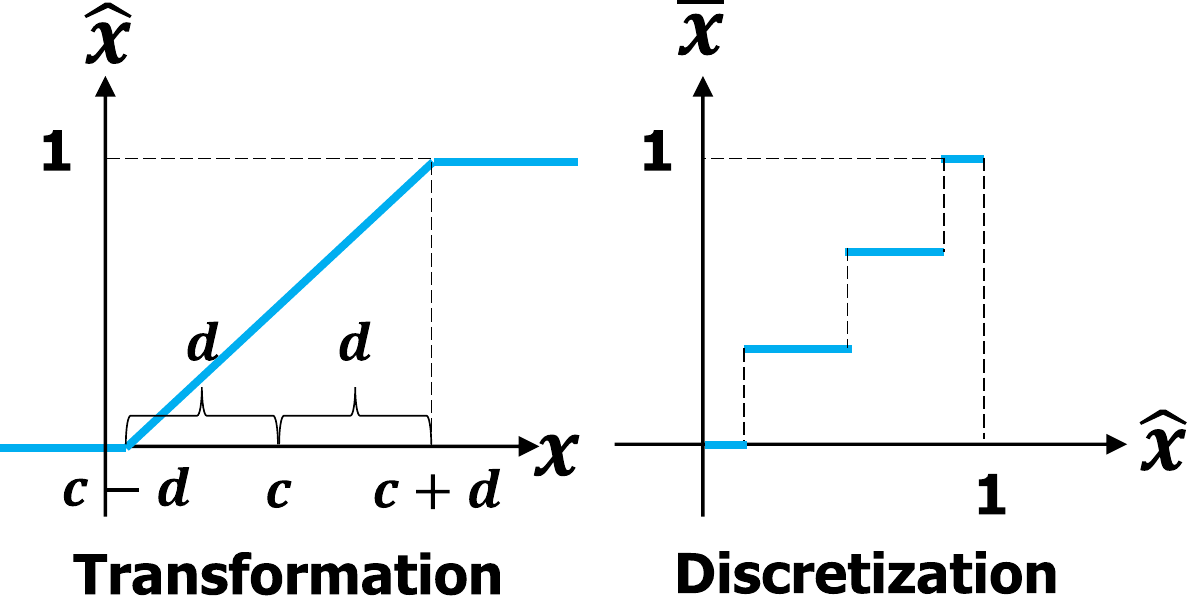}
        \caption{QIL algorithm.} 
        \label{img:qil}
    \end{subfigure}
    \begin{subfigure}{.32\columnwidth}
        \centering
        \includegraphics[width=\columnwidth]{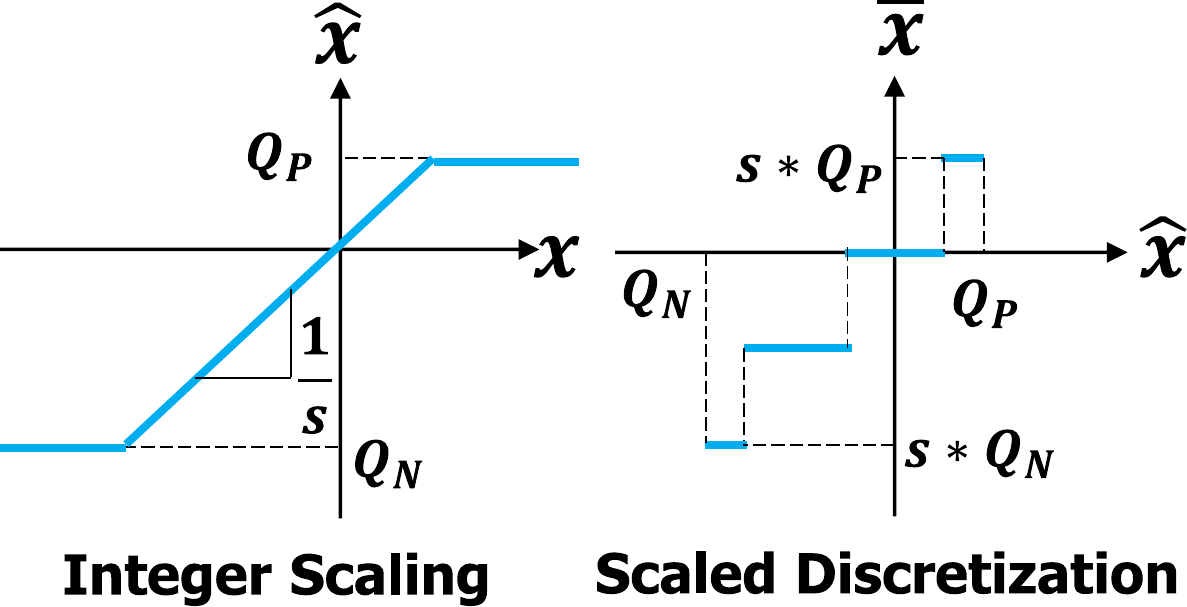}
        \caption{LSQ algorithm.} 
        \label{img:lsq}
    \end{subfigure}
      \\
    \begin{subfigure}{.48\columnwidth}
        \centering
        \includegraphics[width=\columnwidth]{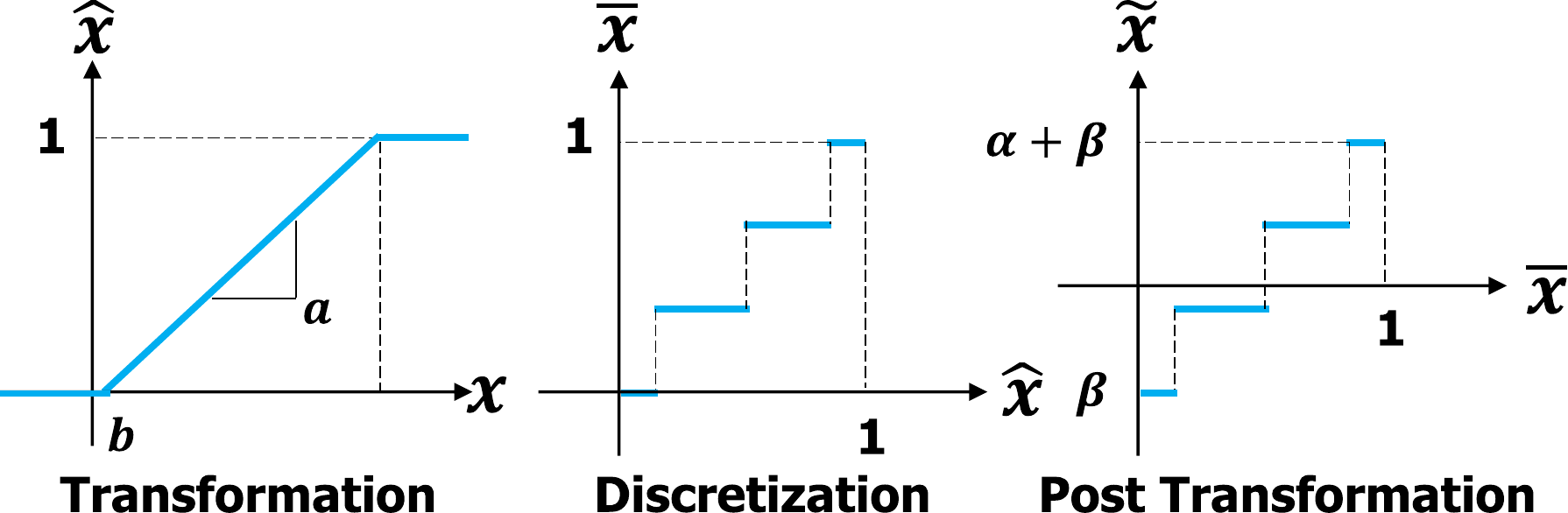}
        \caption{Proposed DuQ algorithm.} 
        \label{img:duq}
    \end{subfigure}
    \begin{subfigure}{.48\columnwidth}
        \centering
        \includegraphics[width=\columnwidth]{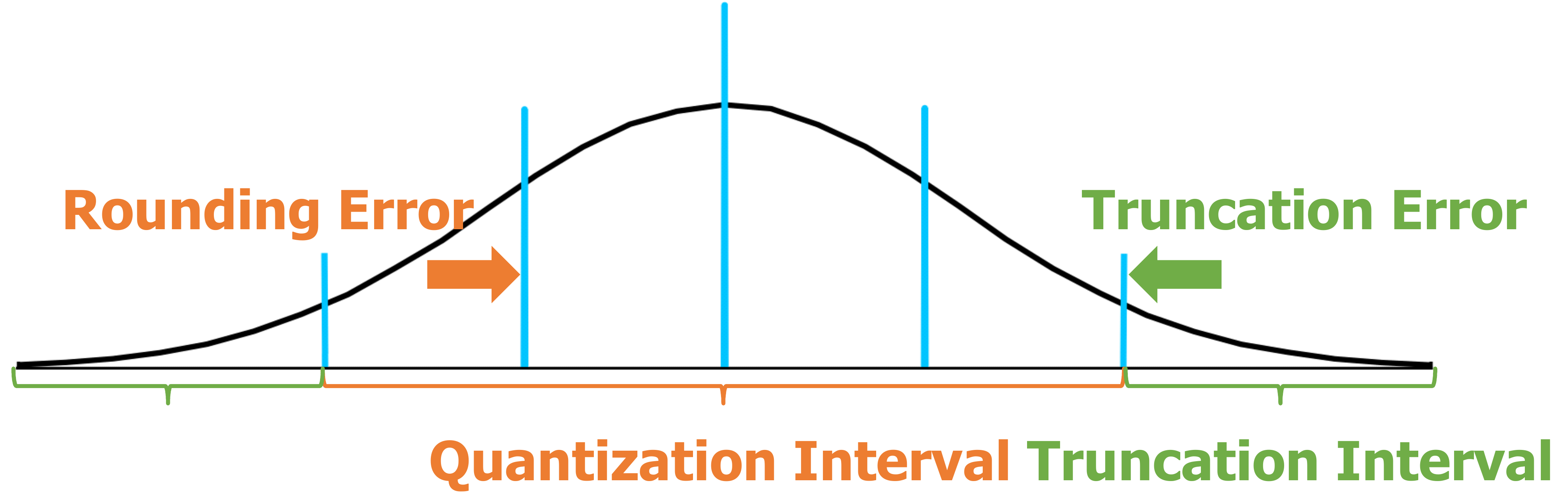}
        \caption{Two error sources of quantization.} 
        \label{img:error}
    \end{subfigure}
    \caption{Quantization algorithm details.}
    \label{img:diffalgo}
\end{figure}

Both PACT and QIL have a critical limitation in supporting new activation functions %, e.g., h-swish and new structures, e.g., squeeze-and-excitation used on state-of-the-art optimized networks. 
because the transformation stage or parameterized clipping stage forces the activation data to be mapped to [0, 1] in \cite{QIL} or [0, $p$] in \cite{PACT}, as shown in Figure~\ref{img:diffalgo} (a) and (b). Thus, it is not applicable to activation functions with asymmetric distributions. %For instance, h-swish activation, squeeze-and-excitation module and linear expansion layer require representing negative activations.
%Since both methods assume only non-negative activations, they do not provide competitive results for optimized networks in the reduced precision like 4 bits.
In the case of LSQ, as shown in Figure~\ref{img:diffalgo} (c), a trainable scale parameter $s$ is adopted, and the hand-crafted parameters utilize the number of negative quantization levels $Q_N$ and that of positive ones $Q_P$. LSQ also handles either symmetric ($Q_N = Q_P = 2^{bit-1}-1$) or non-negative ($Q_N=0, Q_P = 2^{bit}-1$) distributions. Additionally, because the user predetermines the number of quantization levels for negative and positive ranges, it has a limitation in handling various distributions across the layers. %We propose a new quantization algorithm to handle asymmetric distributions considering the layer-wise different distributions. 

\subsection{Proposed Method: DuQ} \label{duq_explain}

Our proposed method, called DuQ, learns the quantization and truncation intervals through back-propagation.% exploiting all the gradients over the entire value range of activation. 
This is an extension of QIL with scale ($\alpha$) and shift($\beta$) parameters that remove the limitation of QIL, i.e., the limited output range of [0, 1]. 

%\begin{figure}[ht!]
%    \centering
%  \includegraphics[width=\columnwidth]{eccv2020kit/figure/duq.pdf}
%  \caption{Proposed DuQ algorithm.} 
%  \label{img:duq}
%\end{figure}
% 은혁, 그림에 a, b 추가

%Our proposed DuQ method improves upon QIL by additionally performing denormalization and utilizing gradients on truncation interval. Thus, we use the same task loss as QIL. 
The two stages of transformation (Eqn. \ref{duq1}) and discretization (Eqn. \ref{duq2}) are identical to those of QIL except that the slope $a$ and offset $b$ are used in the transformation stage instead of the center $c$ and width $d$ in QIL. 
As Eqns. \ref{duq1} and \ref{duq3} show, we use the softplus function for $a$ and $\alpha$ to make them positive values for improving the stability of the transformation stage. Eqn. \ref{duq2} represents the discretization stage, where $N_{lv}$ is the number of quantization levels.

\begin{equation}    
    \hat{x} = clip\big( (x-b)/{a'}, 0, 1\big),~a'=softplus(a), \label{duq1}
\end{equation}
\begin{equation}    
    \bar{x} =  round\big((N_{lv}-1)\cdot\hat{x}\big) / (N_{lv}-1), \label{duq2}
\end{equation}
\begin{equation}    
    \tilde{x} = \alpha'\cdot\bar{x} + \beta,~\alpha'=softplus(\alpha). \label{duq3}
\end{equation}

Our proposed DuQ method allows us to utilize the full value range of activation including the negative ones, and to achieve that, the discretized data can be mapped to an arbitrary range through a scale $\alpha$ and offset $\beta$, in the post-transformation stage, as shown in Eqn. \ref{duq3} as shown in Figure \ref{img:diffalgo} (d). One additional advantage is DuQ utilizes all the gradients across the entire activation data. PACT only utilizes the gradients from the truncation interval (the value range larger than the truncation threshold), while QIL only utilizes the gradients from the quantization interval (between the minimum and maximum quantization levels). Both utilize only a portion of the backpropagated error. However, DuQ utilizes all the gradients to learn a good quantization interval considering the trade-off between rounding and truncation errors, as Figure \ref{img:diffalgo} (e) shows.

\subsection{Negative Padding} \label{negpad}

\begin{figure}[h]
    \centering
  \includegraphics[width=0.85\columnwidth]{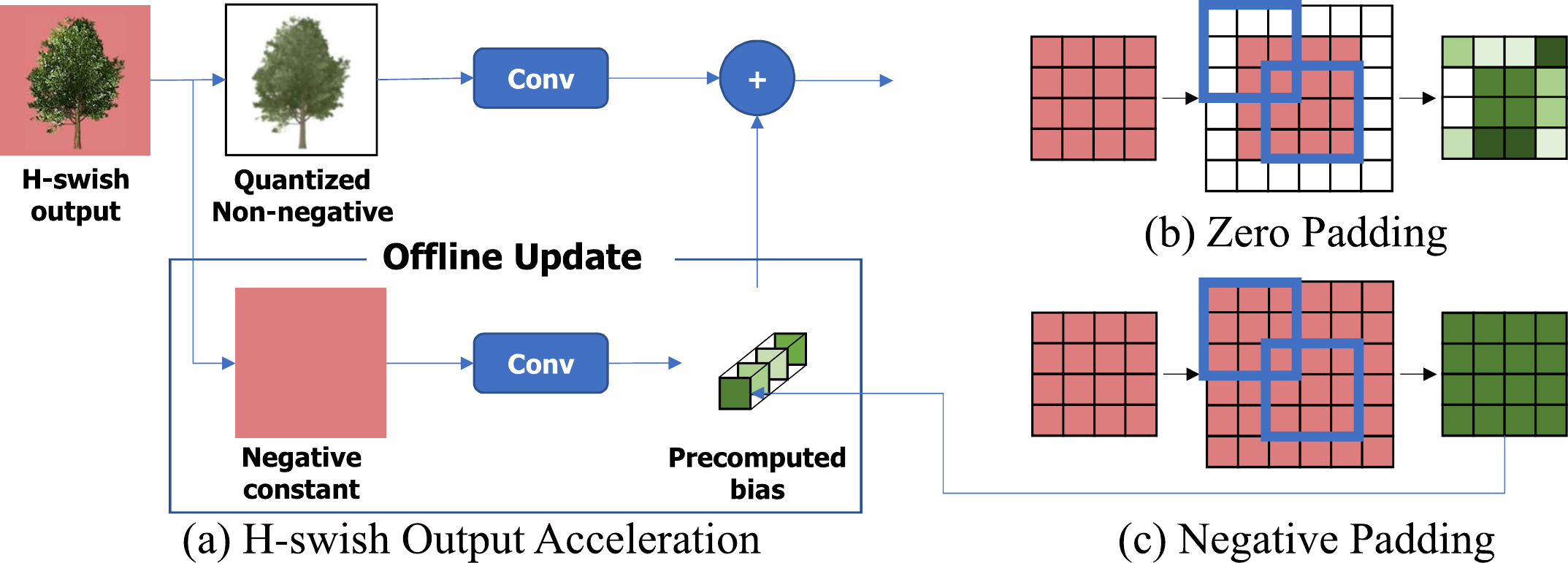}
  \caption{Negative padding for h-swish function.} 
  \label{img:negpad}
\end{figure}

DuQ is flexible enough to support asymmetric distributions. However, many of existing hardware accelerators support only symmetric and non-negative integer types. We propose a idea called negative padding to accelerate the quantized network having asymmetric distributions on such hardware accelerators.
% 은혁, 요즘은 가속기 만들때 범용성 신경을 쓰기 때문에 signed, unsigned 둘 다 지원할꺼야. 그래서 이 문장이 그리 설득력이 없어.
% signed 랑 unsigned 두 가지만 지원하고, non-centered zero를 지원하지 않는다는 점을 말하려 했습니다.

Even though the output of the h-swish function has an asymmetric distribution, it has a constant minimal value, -0.375. As shown in Figure \ref{img:negpad} (a), by shifting the activations by the minimal value, the activations can be broken down into two components, constant negative ones, and shifted non-negative ones. The output corresponding to the constant negative feature can be calculated offline, and the output corresponding to the input-dependent non-negative activations can be efficiently calculated by the hardware accelerator. By combining those two outputs, we can obtain the output. Note that, because of the linearity of the convolution and matrix multiplication, the output is correct.

However, because of conventional zero padding, the output corresponding to constant negative input activations has inconsistent values on the edges, as shown in Figure \ref{img:negpad} (b). We need to store the spatial position-dependent values, which can increase the cost of memory access and computation. This problem can be solved by adopting negative padding instead of zero padding (Figure \ref{img:negpad} (c)). When we pad the edge of the activation with the constant minimal value of the h-swish function, i.e., -0.375, the output has an identical value for all the features in the same spatial dimension. %This constant output can be incorporated into conventional bias addition without additional cost in the test time. 
This constant output enables the pre-computed result to be mapped by the efficient channel-wise bias-add operator, thereby minimizing computational/storage costs. 

The proposed negative padding makes the minimum quantization level of input activation zero. Thus, the proposed method can also be beneficial to zero skipping solutions like zero activation-skipping hardware accelerators \cite{Cnvlutin,SCNN} to improve the inference speed. Please note that the negative padding is applicable to any non-linear activation function, which has asymmetric distributions with a constant minimal value.

\section{Experiments}
We implemented the proposed methods in PyTorch 1.4 and demonstrated their effectiveness by measuring the accuracy of the quantized networks. We applied quantization to well-known optimized CNNs, MobileNet-v1 to v3 and MNasNet-A1. %We also quantize ResNet-18 as a representative example of conventional CNN. 
The networks were trained on 4-GPU with a 256 batch, SGD with momentum, and cosine learning rate decay with warmup~\cite{WARM,SGDR}. In order to improve the accuracy, we adopted the progressive quantization method~\cite{CVPR} that gradually decreases the bit-width to 8, 5, and 4 bits during fine-tuning, and use knowledge distillation~\cite{KD} using the ResNet-101 as teacher. We used an exponential moving average of parameters with a momentum of 0.9997~\cite{EMA}, and all networks were trained using PROFIT and DuQ (with negative padding if applicable). We trained the model for 15 epochs at every progressive quantization and PROFIT fine-tuning step. The entire fine-tuning for the 4-bit network took 140 epochs for weight update. 
Please note that both weights and activations on all the layers of the networks were quantized, including the first and last layers. We did not apply quantization only for the input image of the first convolution layer and the activation of the squeeze-excitation module.

\subsection{Accuracy on ImageNet Dataset}
\begin{table*}[t]
\caption{Top-1 / Top-5 accuracy [\%] of the quantized networks on ImageNet.}
\centering
\small

\begin{tabular}{c|c|c|c|c}
    \toprule
    & \textbf{MobileNet-v1}& \textbf{MobileNet-v2}& \textbf{MobileNet-v3} & \textbf{MNasNet-A1} \\% & \textbf{ResNet-18} \\
    \midrule
    \textbf{Full} & 68.848 / 88.740  & 71.328 / 90.016 & 74.728 / 92.136 & 73.130 / 91.276 \\ %& 69.546 / 89.090 \\
    \textbf{Full+} & 69.552 / 89.138  & 71.944 / 90.470 & 75.296 / 92.446 & 73.396 / 91.464 \\ %& 69.546 / 89.090 \\
    \textbf{8-bit} & 70.164 / 89.370 & 72.352 / 90.636 & 75.166 / 92.498 & 73.742 / 91.756 \\ %& 71.246 / 89.988 \\
    \textbf{5-bit} & 69.866 / 89.058 & 72.192 / 90.498  & 74.690 / 92.092 & 73.378 / 91.244 \\ %& 71.672 / 90.168  \\
    \textbf{4-bit} & 69.056 / 88.412 & 71.564 / 90.398 & 73.812 / 91.588 & 72.244 / 90.584 \\% & 70.968 / 89.914 \\
    
    \bottomrule
\end{tabular}

\label{tab:imagenet}
\end{table*}

In Table \ref{tab:imagenet}, the accuracy of quantized networks under our proposed methods are shown. In the table, 'Full' represents our reproduction of the original training procedure. 'Full+' adopts teacher-student and weight averaging algorithms on top of 'Full'. Compared to the full precision accuracy ('Full+' in the table), our 4-bit models give comparable accuracy at a top-1 accuracy loss of 1.48\%. To the best of the authors' knowledge, this is the SOTA result on the 4-bit quantization of MobileNet-v3 including the first and last layers. From the table, it is also evident that compared with full-precision ('Full+'), our method loses only less than 0.5\% of top-1 accuracy on 4-bit MobileNet-v1 and v2.

\subsection{Comparison with the Existing Works}

%\begin{table}[ht]
%\caption{Top-1 accuracy [\%] comparison of existing %works on MobileNet-v1 and MobileNet-v2.}
%\centering
%\footnotesize

%\begin{tabular}{c|c|c}
%    \toprule 
%    & \textbf{MobileNet-v1}& \textbf{MobileNet-v2} \\
%    \midrule
%    \textbf{\cite{WhitePaper}, a8,w8,c} & 70.7 & 71.1 \\
%    \textbf{\cite{WhitePaper}, a8,w8,l} & 70.0 & 70.9 \\
%    \textbf{\cite{WhitePaper}, a8,w4,c} & 64.0 & 58.0 \\
%    \textbf{\cite{WhitePaper}, a4,w8,c} & 65.0 & 62.0 \\
%    \textbf{\cite{DSQ}, a4, w4} & - & 64.80 \\
%    \cite{FightBias}*, a8,w8 & 69.99 & 70.60 \\
%    \cite{DataFree}*, a8,w8 & 70.5 &  71.2 \\
%    \textbf{Ours, a5,w5} & 69.866 & 72.192 \\
%    \textbf{Ours, a4,w4} & 69.056 & 71.564 \\
%    \bottomrule
%\end{tabular}
%\label{tab:image_comp}
%\end{table}
We compare the accuracy of MobileNet-v1 and v2 with the existing works in Table  \ref{tab:image_comp} where (a,w) represent a-bit activation and w-bit weight quantization and c and l channel-wise and layer-wise quantization, respectively. * represents the post-training quantization. 
Compared to \cite{WhitePaper}, our 4-bit layer-wise quantization gives comparable accuracy to the 8-bit models of previous work, and better accuracy than the channel-wise 8/4-bit models. Furthermore, our 4-bit MobileNet-v2 model outperforms the previous best 4-bit model~\cite{DSQ} by 6.76~\%, and it outperforms even the 8-bit models of existing works~\cite{FightBias,DataFree}.

\subsection{Ablation Study}

\begin{table}[!t]
    \begin{minipage}[t]{0.36\linewidth}
    \centering
    \small
    \caption{Top-1 accuracy [\%] comparison of existing works on MobileNet-v1~(MV1) and MobileNet-v2~(MV2).}
    \label{tab:image_comp}
    \begin{tabular}[t]{l|c|c}
    \toprule 
    & \textbf{MV1}& \textbf{MV2} \\
    \midrule
    \textbf{(8,8), c~\cite{WhitePaper}} & 70.7 & 71.1 \\
    \textbf{(8,8), l~\cite{WhitePaper}} & 70.0 & 70.9 \\
    \textbf{(8,4), c~\cite{WhitePaper}} & 64.0 & 58.0 \\
    \textbf{(4,8), c~\cite{WhitePaper}} & 65.0 & 62.0 \\
    \textbf{(4,4), l~\cite{DSQ}} & - & 64.80 \\
    \textbf{(8,8), l~\cite{FightBias}}* & 70.10 & 70.60 \\
    \textbf{(8,8), l~\cite{DataFree}}* & 70.5 &  71.2 \\
    \textbf{(5,5), l, Ours} & 69.866 & 72.192 \\
    \textbf{(4,4), l, Ours} & 69.056 & 71.564 \\
    \bottomrule
    \end{tabular}
    \end{minipage}
    \hspace{0.8cm}
    \begin{minipage}[t]{0.56\linewidth}
    \caption{Top-1 / Top-5 accuracy [\%] of 4-bit MobileNet-v3 on ImageNet. TS represents teacher-student, PG progressive quantization and PF PROFIT. }
    \centering
    \small
    \label{table:comp}
    \begin{tabular}[t]{c|c|c|c|c}
    \toprule
    \textbf{Algorithm} & \textbf{TS} & \textbf{PG} & 
    \textbf{PF} & \textbf{Accuracy}\\ \midrule
    \textbf{QIL} & & & & fail \\
    \textbf{PACT} & & & & 67.978 / 88.204 \\
    \textbf{PACT} & \checkmark & \checkmark & &  70.160 / 89.576 \\
    \textbf{PACT} & \checkmark & \checkmark & \checkmark & 72.948 / 90.892 \\
    \midrule
    \textbf{DuQ} &  &  &  & 69.504 / 88.946 \\
    \textbf{DuQ} & \checkmark &  &  & 71.006 / 90.018 \\
    \textbf{DuQ} & \checkmark & \checkmark &  & 71.466 / 90.200 \\
    \textbf{DuQ} & \checkmark  & \checkmark & \checkmark  & 73.812 / 91.588 \\
    \textbf{DuQ} &  &  & \checkmark & 72.260 / 90.772 \\
    \textbf{DuQ} & \checkmark  & \checkmark & $\bigtriangleup$  & 72.826 / 91.068 \\
    \midrule
    \textbf{DuQ, Sym}& & & & 68.480 / 88.496 \\
    \textbf{DuQ, Non} & & & & 68.724 / 88.566\\
    \midrule
    \textbf{PACT3b} & \checkmark& \checkmark& & 57.086 / 80.898 \\
    \textbf{PACT3b} & \checkmark & \checkmark& \checkmark&  66.458 / 87.360 \\
    \textbf{DuQ3b} & \checkmark& \checkmark&  &  65.674 / 86.480 \\
    \textbf{DuQ3b} & \checkmark& \checkmark& \checkmark & 69.942  / 89.340 \\
    \bottomrule
    \end{tabular}
    \end{minipage}
\end{table}

% 은혁, PACT 3b 결과가 없어서 abstract/conclusion에서 12.4% 부분 빼야 할 듯. 3b 결과 넣을 수 없나?
% PACT가 너무 낮게 나와서 코드 정리해서 다시 돌리는 중입니다. 지금 trend에서 보면 12.4 %까지는 안나올 것 같습니다. 내일 저녁 즈음 해서 결과가 나올 것 같습니다. 

%\begin{table}[h!]
%\caption{Results of 4-bit MobileNet-v3 on ImageNet. TS represents teacher-student and PG progressive quantization.}
%    \footnotesize
%    \centering
%    \begin{tabular}{ccccc}
%    \toprule
%    \textbf{Algorithm} & \textbf{TS} & \textbf{PG} & 
%    \textbf{PROFIT} & \textbf{Accuracy}\\ \midrule
%    \textbf{QIL} & & & & fail \\
%    \textbf{PACT} & & & & 67.130 / 87.684 \\
%    \textbf{PACT} & \checkmark & \checkmark & & 71.856 / 90.474 \\
%    \textbf{PACT} & \checkmark & \checkmark & \checkmark & 72.948 / 90.892 \\
%    \midrule
%    \textbf{DuQ} &  &  &  & 69.504 / 88.946 \\
%    \textbf{DuQ} & \checkmark &  &  & ? \\
%    \textbf{DuQ} & \checkmark & \checkmark &  & 71.466 / 90.200 \\
%    \textbf{DuQ} & \checkmark  & \checkmark & \checkmark  & 73.812 / 91.588 \\
%    \textbf{DuQ} &  &  & \checkmark & 72.260 / 90.772 \\
%    \textbf{DuQ} & \checkmark  & \checkmark & $\bigtriangleup$  & 72.826 / 90.068 \\

%    \midrule
%    \multicolumn{4}{c}{\textbf{DuQ, symmetric}} & 68.480 / 88.496 \\
%    \multicolumn{4}{c}{\textbf{DuQ, non-negative}} & 68.724 / 88.566\\
%    \midrule
%    \textbf{PACT3b} & \checkmark& \checkmark& & 57.086 / 80.898 \\
%    \textbf{PACT3b} & \checkmark & \checkmark& \checkmark&  66.458 / 87.360 \\
%    \textbf{DuQ3b} & \checkmark& \checkmark& \checkmark & 69.942  / 89.340 \\
%    \bottomrule
%    \end{tabular}
%    \label{table:comp}
%\end{table}

In the ablation study, we compared the existing methods and our proposed method on the 4-bit quantization of MobileNet-v3 on ImageNet. Moreover, to decompose the effect of each of our methods, we evaluated the effect on the existing method and our own method. Note that we use DuQ with negative padding by default except DuQ, Sym (zero-padding with symmetric quantization) and DuQ, Non (zero-padding with non-negative quantization).
We used two existing methods, QIL~\cite{QIL} and PACT with SAWB~\cite{PACT}.
When we applied the QIL algorithm to 4-bit MobileNet-v3, the fine-tuning failed to converge. It is because QIL has a critical limitation that its output range is bounded from 0 to 1, which is detrimental to the squeeze-excitation layer and h-swish activation function. 
As shown in Table~\ref{table:comp}, PACT gives 70.16~\% of top-1 accuracy with a 5.14~\% accuracy drop from full-precision accuracy ('Full+' in Table~\ref{tab:imagenet}). 
Our proposed methods are beneficial to the existing method.
By applying PROFIT to PACT, we can obtain 2.78 \% accuracy improvement, as shown in the table. DuQ outperforms PACT by 1.53~\% (=69.504~\% - 67.978~\%). Under teacher-student and progressive quantization, our solution (DuQ + PROFIT) gives 3.65~\% (=73.812~\% - 70.160~\%) accuracy gain over PACT. 

In order to evaluate the effect of incremental weight freezing, 
we also applied only the last stage of PROFIT that stabilizes the normalization layers by fine-tuning after freezing all the convolution layers. 
This option ($\bigtriangleup$ in the table) gives approximately half the gain of PROFIT, which confirms that PROFIT is essential in minimizing the effect of AIWQ. 

% 은혁, 표에서 DuQ, non-negative와 QIL이 차이 있나?
% QIL은 그냥 수렴에 실패하고, DuQ는 scale 보정해주는 term에 의해서 수렴이 가능합니다. negative padding 유무에 따라 0.780 % 정확도 차이가 존재합니다. 
% Negative padding의 benefits 보여주는 정량적 결과 없으면 논문에서 1페이지나 할애하기 애매함.
The table also shows the effect of negative padding. Comparing DuQ (with negative padding) with DuQ, Sym and DuQ, Non, negative padding gives 1.02~\% and 0.78~\% of accuracy improvement, respectably. In addition, with negative padding, 27.5~\% of activations of h-swish output can be mapped to zero, thus the conventional zero-skipping accelerator can additionally improve performance and energy efficiency without the modification of the network. 
When we reduce the bit-width down to 3-bit (PACT3b and DuQ3b in Table~\ref{table:comp}), our proposed methods bring more accuracy improvement. Compared to the conventional PACT, our methods (DuQ+PROFIT) give 12.86~\% (=69.942~\%-57.086~\%) improvement.
%Our 8-bit model accuracy, 75.166 \% is superior to that (73.8~\%) in \cite{V3}. We think it is mainly because fused-batch normalization, adopted in \cite{V3}, has an adverse effect on the accuracy of low precision model, which will be elaborated in more detail in the next subsection.

The conventional models also suffer from the AIWQ problem in more aggressive quantization, and PROFIT is helpful to improve accuracy. With PROFIT, ResNet-18 can be quantized into 3-bit without accuracy loss and 2-bit with 2.31~\% accuracy loss compared to the baseline model accuracy, which is the state-of-the-art result as far as we know.

\subsection{Computation Cost and Model Size Analysis}

\begin{figure}[t]
    \centering
  \includegraphics[width=\columnwidth]{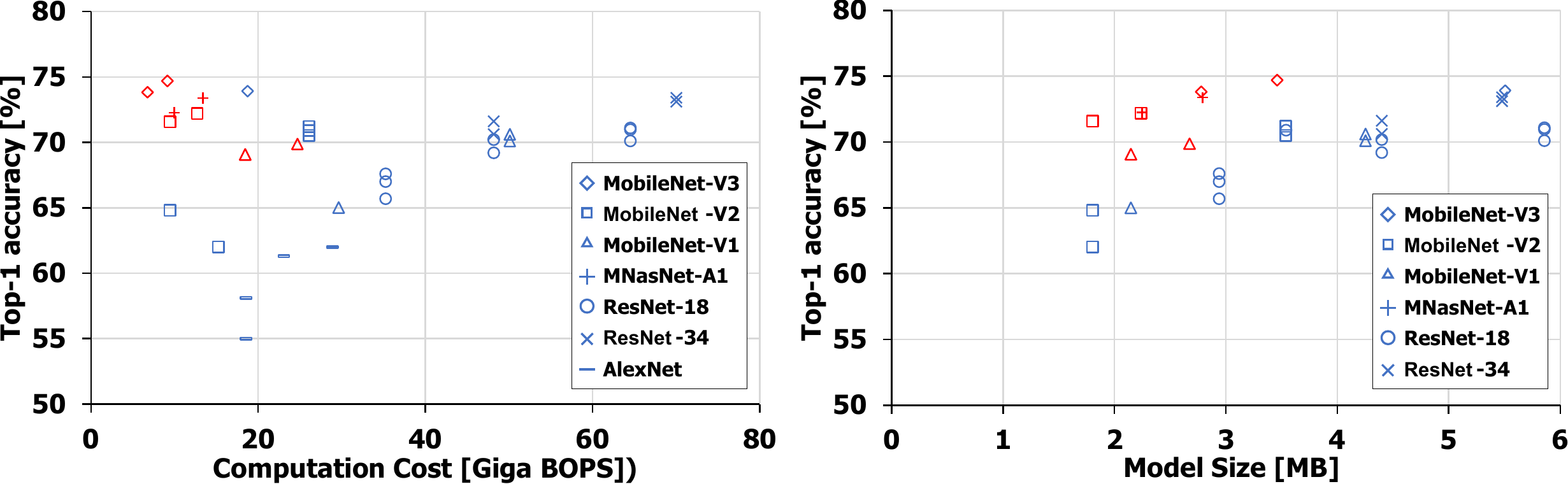}
  \caption{Comparison of accuracy and estimated computation cost based on the HW accelerator model (the bit-operations (BOPS), \cite{UNIQ}), and comparison of accuracy and model size of the quantized network. The tuple (a,w) represents the bit-width of activation and weight, respectively. The red markers represent our results, and the blue markers the results of state-of-the-art methods~\cite{PACT,LSQ,FightBias,DSQ,V3,QIL,WhitePaper,DataFree}.}
  \label{img:perf_model}
\end{figure}

We compared the computation overhead and memory efficiency over the accuracy of the quantized network (Figure \ref{img:perf_model}). We used the bit-operations (BOPS) metric~\cite{UNIQ} that estimates the computation cost based on the required silicon area of the hardware accelerator for the quantized network. In terms of computation, our quantized model offers much higher efficiency with the same accuracy. For instance, at accuracy higher than 73~\%, our 4-bit MobileNet-v3 model takes $2.77\times$ less computation cost than the previous best 8-bit MobileNet-v3 model~\cite{V3}. Compared to 3-bit ResNet-34~\cite{LSQ}, we reduce the cost by up to $10.3\times$. In terms of the model size, our model gives higher accuracy within the same memory constraint. For instance, the 4-bit MobileNet-v2 model shows 6.76~\% higher accuracy than the previous best model (4-bit MobileNet-v2~\cite{DSQ}) within the model size constraint of 2~MB, and the 4-bit MobileNet-v3 model gives 6.21~\% higher accuracy (versus 2-bit ResNet-18~\cite{LSQ}) at 3~MB. These benefits come from our proposed method and the efficiency of the advanced network structure.
% 기존 일들과 비교를 위해 하드웨어 가속기의 면적을 이용해서 연산 오버헤드를 계산하는 metric인 BOP을 도입하고 각 모델 별 model size를 비교하였음. 

% cost: 동일한 연산량 대비 훨씬 높은 성능, 동일 성능에서는 훨신 높은 정확도를 제공함. 예를 들어 73 % 이상의 정확도가 필요한 경우, 우리의 최적화 모델은 기존 MobileNet-v3보다 2.xX 배 더 적은 비용으로 연산이 가능함. ResNet 계열과 비교했을 대에는 XX 배 이상의 연산량을 감소시켰음. 이러한 이득은 대부분 효율적인 네트워크 구조에서 나온 것이나, 정확도를 유지하는 효율적인 양자화를 통해 이를 실현시켰음

% model size의 경우 훨씬 적은 적은 메모리 크기로 높은 정확도 구현 가능. 일례로 2MB 이하의 메모리 한도 내에서 기존보다 x % 높은 정확도 제공. 3 MB 이하에서는 X % 높은 성능. 이 역시 높은 정확도로 양자화가 가능했기 때문에 누릴 수 있는 이득임. 

\section{Conclusion}
We proposed a novel training method called PROFIT, a quantization method called DuQ, and negative padding. PROFIT aims at minimizing the effect of activation instability induced by weight quantization, and DuQ and negative padding enable the quantization of asymmetric distribution in optimized networks. Based on the proposed methods, we can quantize the optimized networks into 4 bits with less than 1.5 \% (MobileNet-v3) and 0.5 \% (MobileNet-v1/2) of top-1 accuracy loss. We anticipate that our proposed methods can contribute to advancing towards sub-4-bit precision computation on mobile and edge devices. 

\section*{Acknowledgement}
This work was supported by Samsung Electronics, the National Research Foundation of Korea grants, NRF-2016M3A7B4909604 and NRF-2016M3C4A7952587 funded by the Ministry of Science, ICT \& Future Planning (PE Class Heterogeneous High Performance Computer Development). We appreciate valuable comments from Dr. Andrew G. Howard and Dr. Jaeyoun Kim at Google.

\clearpage
\bibliographystyle{splncs04}
\bibliography{egbib}
\end{document}